\def\year{2017}\relax
\documentclass[letterpaper]{article}
\usepackage{aaai17}
\usepackage{times}
\usepackage{helvet}
\usepackage{courier}

\usepackage{amssymb}
\usepackage{booktabs}
\usepackage{mathtools}
\usepackage{nth}
\usepackage{graphicx}
\usepackage{subfigure}

\newcommand{\rot}{\rotatebox[origin=c]{90}}

\begin{document}

\title{Training Bit Fully Convolutional Network for Fast Semantic Segmentation}

\author{
\normalsize He Wen \and Shuchang Zhou \and Zhe Liang \and Yuxiang Zhang \and Dieqiao Feng \and Xinyu Zhou \and Cong Yao \\
Megvii Inc. \\
{\{wenhe, zsc, liangzhe, zyx, fdq, zxy, yaocong\}@megvii.com}
}

\maketitle

\begin{abstract}
Fully convolutional neural networks give accurate, per-pixel prediction for input images and have applications like semantic segmentation. However, a typical FCN usually requires lots of floating point computation and large run-time memory, which effectively limits its usability. We propose a method to train Bit Fully Convolution Network (BFCN), a fully convolutional neural network that has low bit-width weights and activations. Because most of its computation-intensive convolutions are accomplished between low bit-width numbers, a BFCN can be accelerated by an efficient bit-convolution implementation.
On CPU, the dot product operation between two bit vectors can be reduced to bitwise operations and popcounts, which can offer much higher throughput than 32-bit multiplications and additions.

To validate the effectiveness of BFCN, we conduct experiments on the PASCAL VOC 2012 semantic segmentation task and Cityscapes. Our BFCN with 1-bit weights and 2-bit activations, which runs 7.8x faster on CPU or requires less than 1\% resources on FPGA, can achieve comparable performance as the 32-bit counterpart.
\end{abstract}

\section{Introduction}
Deep convolutional neural networks (DCNN), with its recent progress, has
considerably changed the landscape of computer vision \cite{krizhevsky2012imagenet} and many other fields.

To achieve close to state-of-the-art performance, a DCNN usually has a lot of
parameters and high computational complexity, which may easily overwhelm
resource capability of embedded devices. Substantial research
efforts have been invested in speeding up DCNNs on both general-purpose
\cite{vanhoucke2011improving,gong2014compressing,han2015learning} and
specialized computer hardware  \cite{farabet2009cnp,farabet2011large,pham2012neuflow,chen2014diannao,chen2014dadiannao,zhang2015optimizing}.

\begin{table}[!ht] \centering
\begin{center}
\begin{tabular}{c c c c}
	\toprule \textbf{network}	& \textbf{VOC12} & \textbf{Cityscapes} & \textbf{speedup}
    \\
    \midrule 32-bit FCN		& 69.8\%	& 62.1\%		& 1x
    \\
    \midrule 2-bit BFCN		& 67.0\%	& 60.3\%		& 4.1x
    \\
    \midrule 1-2 BFCN		& 62.8\%	& 57.4\%		& 7.8x
    \\
\hline
\end{tabular}
\end{center}
\caption{Summary results of our BFCNs. Performance measure in mean IoU.}
\label{tab:summary_result}
\end{table}

Recent progress in using low bit-width networks has
considerably reduced parameter storage size and computation burden
by using 1-bit weight and low bit-width activations.
In particular, in BNN \cite{kim2016bitwise} and XNOR-net \cite{rastegari2016xnor}, during the forward pass the
most computationally expensive convolutions can be done by combining xnor
and popcount operations, thanks to the following equivalence when $x$ and $y$
are bit vectors:

\begin{align*}
\sum_i^n x_i y_i = n - 2\operatorname{popcount}(\operatorname{xnor}(x_i, y_i))
\text{, } x_i, y_i \in \{-1, 1\},  \forall i \text{.}
\end{align*}

Specifically, an FPGA implementation of neural network can take more benefit from low bit-width computation,
because the complexity of a multiplier is proportional to the square of bit-widths.

However, most of previous researches on low bit-width networks have been focused on
classification networks. In this paper, we are concerned with
fully convolutional networks (FCN), which can be thought of as performing
pixelwise classification of the input images and have applications in tasks like
semantic segmentation \cite{long2015fully}.
Techniques developed in this paper can also be applied to other variants like RPN \cite{ren2015faster},
FCLN \cite{johnson2015densecap} and Densebox \cite{huang2015densebox}.
Compared to a typical classification network, the following properties of FCN make it a better candidate to apply low bit-width quantizations.

\begin{enumerate}
  \item An FCN typically has large feature maps, and some of them may need to be
  stored for later combination, which pushes up its peak memory usage. As
  BFCN uses low bit-width feature maps, the peak memory usage is significantly reduced.
  
  \item An FCN usually accepts large input image
  and taps into a powerful classification network like VGGNet \cite{Simonyan14c}
  or ResNet \cite{he2015deep} to boost performance. The acceleration offered by
  exploiting bit-convolution kernel, together with memory savings,
  would allow BFCN to be run on devices with limited computation resources.
\end{enumerate}

Considering the method of training a low bit-wdith network is still under exploration, it remains a challenge to
find a way to train a BFCN efficiently as well.

Our paper makes the following contributions:
\begin{enumerate}
  \item We propose BFCN, an FCN that has low bit-width weights and activations,
  which is an extension to the combination of
  methods from Binarized Neural Network \cite{courbariaux2014training}, XNOR-net
  \cite{rastegari2016xnor} and DoReFa-net \cite{zhou2016dorefa}.
  
  \item We replace the convolutional filter in reconstruction with residual blocks
  to better suit the need of low bit-width network. We also propose a novel
  bit-width decay method to train BFCN with better performance.
  In our experiment, 2-bit BFCN with residual reconstruction and linear bit-width decay
  achieves a 67.0\% mean intersection-over-union score, which is 7.4\% better than the vanilla variant.
  
  \item Based on an ImageNet pretrained ResNet-50 with bounded weights and activations,
  we train a semantic segmentation network with 2-bit weights and activations except for the first layer,
  and achieves a mean IoU score of 67.0\% on PASCAL VOC 2012 \cite{Everingham15} and 60.3\% on Cityscapes \cite{Cordts2016Cityscapes}, both on validation set as shown in \ref{tab:summary_result}.
  For comparison, the baseline full-precision model is 69.8\% and 62.1\% respectively.
  Our network can run at 5x speed on CPU compared to full-precision,
  and can be implemented on FPGA with only few percents resource consumption.
  
\end{enumerate}

\section{Related Work}
Semantic segmentation helps computer to understand the structure of images,
and usually serves as a basis of other computer vision applications.
Recent state-of-the-art networks for semantic segmentation are mostly fully convolutional networks \cite{long2015fully} and adopt the architecture of encoder-decoder
with multi-stage refinement \cite{badrinarayanan2015segnet}.
In order to achieve best performance, powerful classification models are
often embedded as part of the FCNs, which pushes up computational complexity together with large decoders.

To further refine the results from neural networks, CRFs are widely used in post-processing to improve local predictions \cite{chen2014semantic}
by reconstructing boundaries more accurately.
Since CRF can be integrated with most methods as post-processing step, which contributes little to our main topic, it will not be discussed in this paper.

Recent success of residual network has shown that very deep networks
can be trained efficiently and performs better than any other previous network.
There also exists successful attempt \cite{wu2016high} to combine FCN and ResNet,
which achieves considerable improvement in semantic segmentation.

To utilize scene-parsing network in low-latency or real-time application, the computational complexity need to be significantly reduced.
Some methods \cite{paszke2016enet,kim2016pvanet} are proposed to reduce demand of computation resources of FCN
by simplifying or redesigning the architecture of network.

We also note that our low bit-width method can be integrated with almost all other speed-up methods and achieves
even further acceleration. For example, low-rank
approaches \cite{zhang2015accelerating} is orthogonal to our approach and may be integrated to BFCN.

\section{Method}
In this section we first introduce the design of our bit fully convolutional network,
and then propose our method for training a BFCN.

\subsection{Network design}

\begin{figure*}
\begin{center}
\includegraphics[scale=0.85]{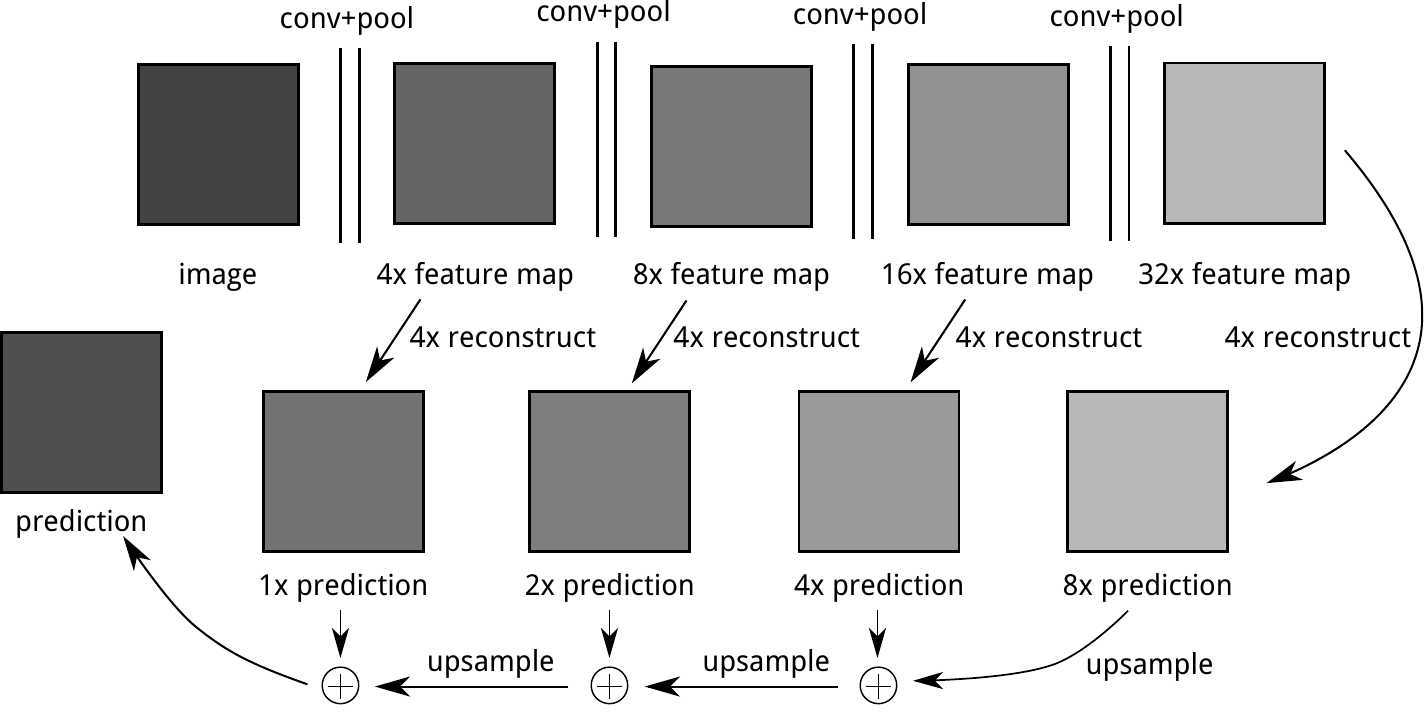}
\end{center}
   \caption{Network Architecture.}
\label{fig:net_arch}
\end{figure*}

A standard approach to perform semantic segmentation includes a feature extractor
to produce feature maps from input image, and convolutions with upsampling operations to
predict per-pixel labels from those feature maps.
We use ResNet as feature extractor and adopt the multi-resolution reconstruction structure from Laplacian Reconstruction and Refinement \cite{ghiasi16lrr}
to perform per-pixel classification over feature maps in different scales (see Figure \ref{fig:net_arch}).

However, while the network works well in full-precision, we observe a great loss in accuracy
while converting it to low bit-width, indicating that this architecture is not suitable for low bit-width network.
To address this issue, we evaluate different variations of BFCN, so as to figure out the cause of performance degeneration.
As shown in Table \ref{tab:bfcn_var}, low bit-width network with single convolution in reconstruction structure
suffer great performance degeneration.
We also discover that adding more channels in reconstruction filter helps improve performance considerably, indicating 
a low bit-width convolution is not enough to extract spatial context from feature maps.
In short, we need a more low bit-width friendly architecture in reconstruction to eliminate the bottleneck.

\begin{table}[!ht] \centering
\begin{center}
\begin{tabular}{c c c}
	\toprule \textbf{model}			& \textbf{mean IoU}	& \textbf{Ops in reconstruction}
    \\
    \midrule 32-bit model			& 66.4\%	& 3.7 GOps
    \\
    \hline
    \midrule baseline				& 59.6\%	& 3.7 GOps
    \\
    \midrule 2x filter channel		& 63.3\%	& 7.8 GOps
    \\
    \midrule residual block			& 67.0\%	& 6.9 GOps
    \\
\hline
\end{tabular}
\end{center}
\caption{Comparison of different variations of low bit-width reconstruction structures on PASCAL VOC 2012. All variants except "residual block" use single convolution as filter in reconstruction.}
\label{tab:bfcn_var}
\end{table}

Intuitively, we may add more channels to the filter in reconstruction.
But it also pushes up computational complexity a lot.
Fortunately, ResNet \cite{he2015deep} allows us to go deeper instead of wider.
It has been shown that a deeper residual block can often performs better than a wider convolution.
Therefore, we address the issue by replacing the linear convolution with residual blocks.
As shown in Table \ref{tab:bfcn_var}, our residual block variant even outperforms the original full-precision network.

In our approach, residual reconstruction structure can not only achieve better performance
with similar complexity to a wide convolution,
but also accelerate training by reduce the length of shortest path in reconstruction.

\subsection{Bit-width allocation}
It is important to decide how many bits to allocate for weights and feature maps,
because bit-width has a crucial impact on both performance and computational complexity of a network.
Since our goal is to speed-up semantic segmentation networks without losing much performance,
we need to allocate bit-width carefully and wisely.

First we note it has been observed \cite{gupta2015deep} that 8-bit fixed point quantization is enough for a network
to achieve almost the same performance as 32-bit floating point counterpart.
Therefore, we focus our attention to bit-widths less than eight, which can provide us with further acceleration.

In order to extend bit-convolution kernel to $m$-bit weights and $n$-bit feature maps, we notice that:
\begin{align*}
	A \times B =& (A_0+2A_1+...+2^mA_m)(B_0+...+2^nB_n) \\
		=& A_0B_0+...+2^{i+j}A_iB_j+...+2^{m+n}A_mB_n
\end{align*}
where $A_i$, $B_i$ represent the i-th bit of $A$ and $B$.
Therefore, it is pretty straightforward to compute the dot product using
$m \cdot n$ bit-convolution kernels for $m$-bit weights and $n$-bit feature maps.
It shows that the complexity of bit-width allocation, which is our primary goal to optimize,
is proportional to the product of bit-widths allocated to weights and activations.
Specifically, bit-width allocation becomes vital on FPGA implementation since it is the direct restriction of network size.

With fixed product of bit-widths, we still need to allocate bits between weights and activations.
Intuitively, we would allocate bits equally as it keeps a balance between weights and activations, and error analysis confirms this intuition.

We first note that the error of a number introduced by $k$-bit quantization is $1/2^k$.
As the errors are accumulated mainly by multiplication in convolution, it can be estimated as follow:

\begin{align}
	\label{eq:error}
	E = \frac{1}{2^{k_W}} + \frac{1}{2^{k_A}}
	  = \frac{2^{k_W}+2^{k_W}}{2^{k_W+k_A}}
\end{align}

When $c=k_W \cdot k_A$ is constant, we have the following inequality:

\begin{align}
	\label{eq:ieq_error}
	E \ge \frac{2\times 2^{\sqrt{c}}}{2^{2\sqrt{c}}}
\end{align}

The equality holds iff $k_W=k_A=\sqrt{c}$, thus a balanced bit-width allocation is needed so as to minimize errors.

For the first layer, since the input image is 8-bit, we also fix bit-width of weights to 8.
The bit-width of activations is still the same as other layers.


\subsection{Route to low bit-width}

\begin{figure}
\begin{center}
\includegraphics[scale=0.45]{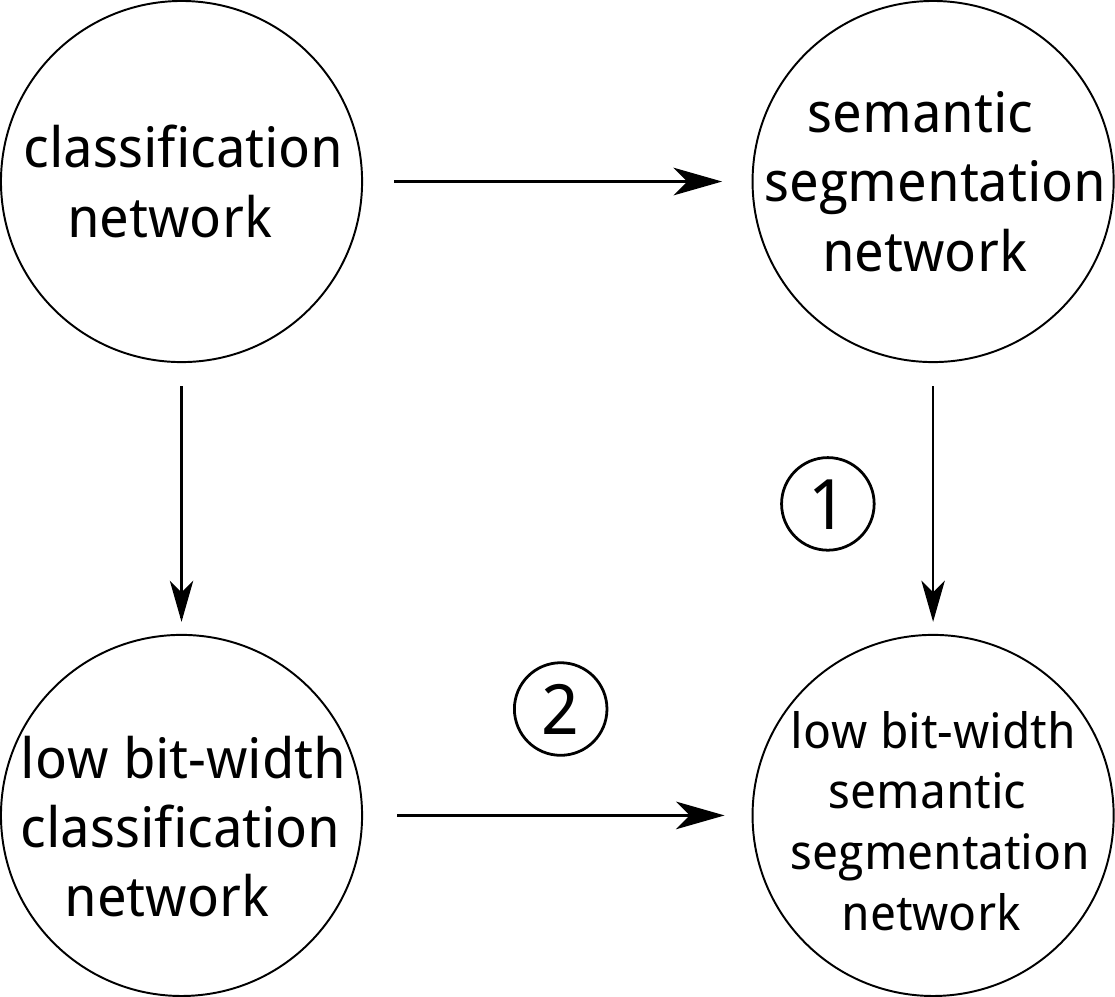}
\end{center}
   \caption{Routes of training BFCN.}
\label{fig:routes}
\end{figure}

\begin{table}[!ht] \centering \small
\begin{center}
\begin{tabular}{c c}
	\toprule initialization			& mean IoU
    \\
    \midrule low bit-width ResNet	& 63.5\%
    \\
    \midrule 32-bit FCN				& 65.7\%
    \\
    \midrule 8-bit BFCN				& 65.8\%
    \\
\hline
\end{tabular}
\end{center}
\caption{Results of different routes of training 2-bit BFCN on PASCAL VOC 2012 val set.}
\label{tab:routes}
\end{table}

As shown in Figure \ref{fig:routes}, there are two ways to adapt the procedure of training
a full-precision fully convolutional network
to produce BFCN, denote as \texttt{P1} and \texttt{P2}.

The only difference between \texttt{P1} and \texttt{P2} is the initialization.
\texttt{P1} uses full-precision FCN as initialization while \texttt{P2} uses low bit-width classification network.
Here full-precision FCN serves as a intermediate stage in the procedure of training.

We evaluate the two routes and find the former one performs significantly better as the mean IoU scores indicate in Table \ref{tab:routes}.
We then add one more intermediate stage to the procedure, the 8-bit BFCN, and achieve a slightly better result.
We conjecture that utilizing intermediate network helps to preserve more information in the process of converting to low bit-width.

\subsection{Bit-width decay}

\begin{figure}
\begin{center}
\includegraphics[height=0.3\textheight, width=0.48\textwidth]{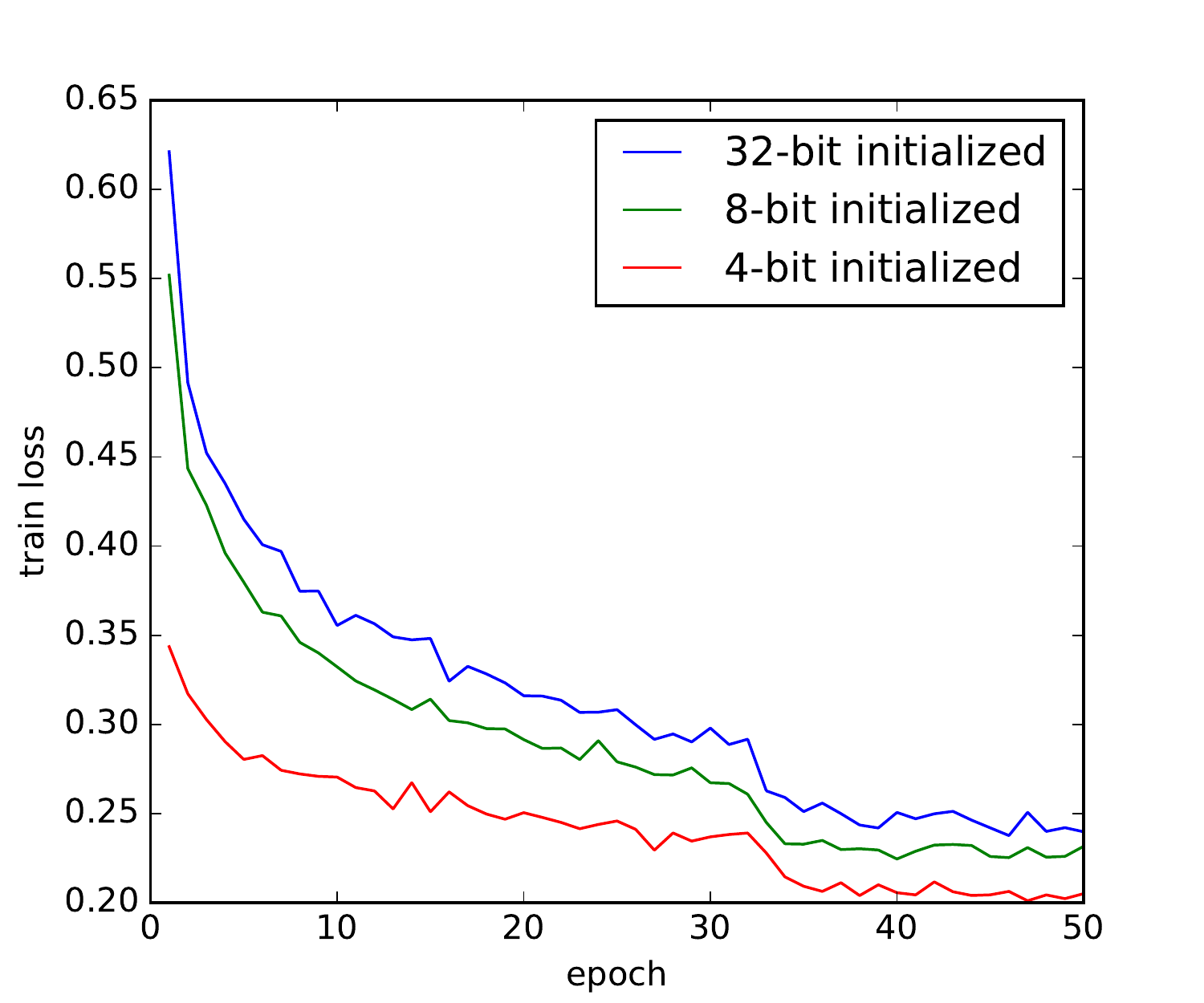}
\end{center}
   \caption{Training loss with different initializations of 2-bit BFCN. Experiments are conducted on PASCAL VOC 2012. The 4-bit model is initialized from 8-bit model.}
\label{fig:train_loss}
\end{figure}

We notice that cutting off bit-width directly from full-precision to very low bit-width
will lead to significant performance drop.
To support this observation, we perform a simple experiment by training a 2-bit network initialized by a pretrained network of different number of bits.
The training process (Figure \ref{fig:train_loss}) shows that networks initialized from lower bit-width converge faster.

This phenomenon can be explained by looking at the errors in quantization.
Obviously, with higher original precision, a quantization step introduced larger error,
and as as result the model benefit less from the initialization.
However, introducing intermediate stages can help resolve it since networks with closer bit-widths tend to be more similar, hence more noise-tolerant when cutting off bit-width.

Our experiments show that BFCN can not recover from the quantization loss very well, if directly initialized from full-precision models.
To extend the idea of utilizing intermediate models during
training low bit-width network, we add more intermediate steps to train BFCN.
We propose a method called bit-width decay, which cuts
off bit-width step-by-step to avoid the overwhelming quantization error caused by large numeric precision drop.

We detail the procedure of bit-width decay method as follow:

\begin{enumerate}
	\item Pretrain a full-precision network $N_1$.
	\item Quantize $N_1$ to produce $N_2$ in 8-bit, which has been proved to be lossless, and fine-tune until its convergence.
	\item Initialize $N_3$ with $N_2$.
	\item Decrease bit-width of $N_3$, and fine-tune for enough iterations.
	\item Repeat step~4 until desired bit-width is reached.
\end{enumerate}

In this way, we can reduce the unrecoverable loss of quantization and the adverse impact of quantization can be mostly eliminated.

\section{Experiments}
In this section, we first describe the datasets we evaluate on and the experiment setup,
then demonstrate the results of our method.
Note that we conduct most of our experiments in our in-house machine learning
system. 

\begin{table*}[!ht] \centering \small
\begin{center}
\tabcolsep=0.10cm
\begin{tabular}{cccccccccccccccccccccc}
	\toprule method	& \rot{mean} & \rot{aero} & \rot{bike} & \rot{bird} & \rot{boat} &
	 	\rot{bottle} & \rot{bus} & \rot{car} & \rot{cat} & \rot{chair} & \rot{cow} &
	 	\rot{table} & \rot{dog} & \rot{horse} & \rot{motor} & \rot{person} & 
	 	\rot{plant} & \rot{sheep} & \rot{sofa} & \rot{train} & \rot{tv}
    \\\hline
    \midrule 32-bit FCN & \textbf{69.8}	& \textbf{82.4} & 38.9 & \textbf{82.7} & 64.3 & 66.4 & \textbf{86.9} & \textbf{83.3} & \textbf{86.5} & \textbf{31.5} & \textbf{73.0} & 48.9 & \textbf{78.6} & \textbf{65.4} & \textbf{77.3} & 81.0 & \textbf{55.7} & 79.3 & 41.3 & 77.8 & 65.1
    \\
    \midrule 4-bit BFCN & 68.6	& 82.3 & 39.1 & 79.4 & \textbf{67.4} & \textbf{66.5} & 85.9 & 79.6 & 84.9 & 29.9 & 69.3 & \textbf{50.1} & 75.9 & 63.2 & 74.9 & \textbf{81.1} & 55.6 & \textbf{82.3} & 37.5 & \textbf{78.0} & \textbf{66.2}
    \\
    \midrule 2-bit BFCN & 67.0	& 80.8 & \textbf{39.7} & 75.4 & 59.0 & 63.2 &
    					85.2 & 79.5 & 83.6 & 29.7 & 71.3 & 44.2 & 75.6 &
    					63.8 & 73.1 & 79.7 & 48.5 & 79.6 & \textbf{43.4} & 74.4 & 65.4
    \\
\hline
\end{tabular}
\end{center}
\caption{Class-wise results on PASCAL VOC 2012 val set.}
\label{tab:pascal_voc_detail_results}
\end{table*}

\subsection{Datasets}
We benchmarked the performance of our BFCN on PASCAL VOC 2012 and Cityscapes, two popular datasets for semantic segmentation.

The PASCAL VOC 2012 dataset on semantic segmentation consists of 1464 labelled images for
training, and 1449 for validation. There are 20 categories to be predicted, including
aeroplane, bus, chair, sofa, etc. All images in the dataset are not larger than 500x500.
Following the convention of literature \cite{long2015fully,wu2016high}, we use the augmented dataset from \cite{hariharan2011semantic},
which gives us 10582 images for training in total. We also utilized reflection, resizing
and random crop to augment the training data.

The Cityscapes dataset consists of 2975 street photos with fine annotation for training and 500 for validation.
There are 19 classes of 7 categories in total.
All images are in resolution of 2048x1536.
In our experiment, the input of BFCN is random-cropped to 1536x768 due to GPU memory restriction,
while validation is performed in its original size.
We train our models with fine-annotated images only.

For performance evaluation, we report the mean class-wise intersection-over-union score (mean IoU),
which is the mean of IoU scores among classes.

\subsection{Experiment Setup}

All experiments are initialized from a ImageNet pretrained ResNet-50 with bounded activations and weights.
We then use stochastic gradient descend with momentum of 0.9 to
fine-tune the BFCN on semantic segmentation dataset.

Since the prediction on higher resolution feature maps in laplacian reconstruction and refinement structure depends on
the prediction on lower resolutions, we use stage-wise losses to train the network.
At first, we only define loss on 32x upsampling branch and fine-tune the network until convergence.
Then losses of 16x, 8x and 4x upsampling branches are added one by one.

In order to overcome the huge imbalance of classes in Cityscapes dataset, we utilize a class weighing scheme introduced by ENet,
which is defined as $W_{class}=1/\ln(c+p_{class})$. We choose $c=1.4$ to bound class weights in $[1, 3]$.

\subsection{Results of different bit-width allocations}

\begin{table}[!ht] \centering \small
\begin{center}
\begin{tabular}{c c c}
	\toprule \textbf{bit-width (W / A)} & \textbf{mean IoU} & \textbf{Complexity}
    \\
    \midrule 32 / 32 & 69.8\%	& -
    \\
    \midrule 8 / 8 & 69.8\%		& 64
    \\
    \midrule 4 / 4 & 68.6\%		& 16
    \\
    \midrule 3 / 3 & 67.4\%		& 9
    \\
    \midrule 2 / 2 & 65.7\%		& 4
    \\
    \midrule 1 / 4 & 64.4\%		& 4
    \\
    \midrule 4 / 1 & diverge	& 4
    \\
    \midrule 1 / 2 & 62.8\%		& 2
    \\
\hline
\end{tabular}
\end{center}
\caption{Results of different bit-width allocated to weight and activation on PASCAL VOC 2012 val set. W represents weight and A for activation. Complexities are measured in terms of the number of bit-convolution kernels needed to compute low bit-width convolution.}
\label{tab:bit-width_results}
\end{table}

First we evaluate the impact of different bit-width allocations on PASCAL VOC 2012 dataset
(see Table \ref{tab:bit-width_results}).

We observe the performance of network degenerates while bit-width is decreasing, which correspond to our intuition.
While 8-8 model performs exactly the same as the full-precision model,
decreasing bit-width from 4-4 to 2-2 continuously incurs degeneration in performance.
The performance degeneration is at first minor compared to bit-width saving, but suddenly becomes non-negligible around 4-4.
We also discover that allocating different bit-widths to weights and activations harms performance compared to equally-allocated model with the same complexity.

From the results we conclude that 4-4 and 2-2 are favorable choices in different scenes.
The 4-4 model can offer comparable performance with full-precision model
but with considerable 75\% resource savings compared to 8-8 on FPGA.
In a more resource-limited situation, the 2-2 model can still offer good performance with only 6.25\% hardware complexity of 8-8 model.

\subsection{Results of bit-width decay}

\begin{table}[!ht] \centering
\begin{center}
\begin{tabular}{c c}
	\toprule \textbf{decay rate}		& \textbf{mean IoU}
    \\
    \midrule 1				& 67.0\%
    \\
    \midrule 2				& 66.8\%
    \\
    \midrule 3				& 66.1\%
	\\
	\hline
	\midrule no decay		& 65.8\%
	\\
\hline
\end{tabular}
\end{center}
\caption{Results of different decay rates.}
\label{tab:decay_rate}
\end{table}

We then show how bit-width decay affects performance of networks on PASCAL VOC 2012.

It can be seen from Table \ref{tab:decay_rate} that bit-width decay does help to achieve
a better performance compared to directly cutting off bit-width.

Besides, we evaluate the impact of "decay rate", which is the number of bits in a step.
For a decay rate of $r$, we have $k_W = c - r \cdot t$ and $k_A = c - r \cdot t$ after $t$ steps of decay, where $c=8$ is the initial bit-width.
The results of different decay rates are also presented in Table \ref{tab:decay_rate}.

We discover with decay rate less than 2 we can achieve almost the same performance,
but increasing it to 3 leads to a sudden drop in performance.
It indicates network with 3 less bits starts diverging from its high bit-width couterpart.

\begin{table*}[!ht] \centering \small
\begin{center}
\tabcolsep=0.12cm
\begin{tabular}{cccccccccccccccccccccc}
	\toprule method	& \rot{mean} & \rot{road} & \rot{sidewalk} & \rot{building} & \rot{wall} &
	 	\rot{fence} & \rot{pole} & \rot{light} & \rot{sign} & \rot{veg} & \rot{terrain} &
	 	\rot{sky} & \rot{person} & \rot{rider} & \rot{car} & \rot{truck} & 
	 	\rot{bus} & \rot{train} & \rot{moto} & \rot{bicycle}
    \\\hline
    \midrule 32-bit FCN & \textbf{62.1}	& \textbf{95.8} & \textbf{73.5} & \textbf{88.2} & \textbf{31.4} & \textbf{38.2} & \textbf{52.6} & \textbf{49.6} & \textbf{65.8} & \textbf{89.8} & \textbf{52.7} & \textbf{90.1} & \textbf{72.8} & \textbf{47.6} & \textbf{89.9} & \textbf{40.8} & \textbf{57.3} & \textbf{37.2} & \textbf{38.2} & \textbf{69.1}
    \\
    \midrule 2-bit BFCN & 60.3	& 95.3 & 71.2 & 87.6 & 25.9 & 36.2 & 51.8 & 49.0 & 63.3 & 89.1 & 51.7 & 89.9 & 71.4 & 44.5 & \textbf{89.9} & 40.1 & 53.8 & 32.4 & 33.8 & 67.5
    \\
    \midrule 1-2 BFCN 	& 57.4	& 94.4 & 70.1 & 86.5 & 22.7 & 33.9 & 49.9 & 44.3 & 62.2 & 87.9 & 44.9 & 89.3 & 69.5 & 40.0 & 88.1 & 35.1 & 51.8 & 21.0 & 32.6 & 65.9
    \\
\hline
\end{tabular}
\end{center}
\caption{Class-wise results on Cityscapes val set.}
\label{tab:cityscapes_detail_results}
\end{table*}

\subsection{Analysis of class-wise results}

We demonstrate our class-wise results of PASCAL VOC 2012 and Cityscapes in Table \ref{tab:pascal_voc_detail_results} and \ref{tab:cityscapes_detail_results}.

As can be observed that most performance degeneration occur in classes which are more difficult to classify.
In PASCAL VOC 2012, we observe that on fine-grained classes like car and bus, cat and dog, BFCN is less powerful than its 32-bit counterpart, however on classes like sofa and bike, 2-bit BFCN even outperforms the full-precision network.

It can be seen more clearly on Cityscapes dataset: classes
with low mean IoU scores in full-precision network become
worse after quantization (like wall and train), while those
large-scale, frequent classes such as sky and car
remain in nearly the same accuracy.

The observation correspond to our intuition that a low bit-width quantized network is usually less powerful and
thus harder to train on difficult tasks.
It also suggest that we may use class balancing or bootstrapping to improve performance in these cases.

\subsection{Analysis of run-time performance}

We then analyze the run-time performance of BFCN on Tegra K1's CPU.
We have implemented a custom runtime on arm, and all our results on CPU are measured directly in the runtime.

We note that 1 single precision operation is equivalent to 1024 bitOps on FPGA
in terms of resource consumption, and roughly 18 bitOps on CPU according to the inference  speed measured in our custom runtime.
Thus, a network with $m$-bit weights and $n$-bit activations is expected to be
$\frac{18}{m \cdot n}$ faster than its 32-bit counterpart ignoring the overheads.

As shown in Table \ref{tab:perf_results}, our 1-2 BFCN can run 7.8x faster than full-precision network with only 1/32 storage size.

\begin{table}[!ht] \centering
\begin{center}
\begin{tabular}{c c c c}
	\toprule \textbf{method}	& 	\textbf{run time} 		& \textbf{parameter size}
	\\
    \midrule 32-bit FCN	& 9681 ms	& 137.7 MB
    \\
    \midrule 1-2 BFCN & 1237 ms	& 4.3 MB
    \\
\hline
\end{tabular}
\end{center}
\caption{Comparison in hardware requirements of Cityscapes models. Run times are measured on Tegra K1 using single core for input size of 256x160x3.}
\label{tab:perf_results}
\end{table}

\subsection{Discussion}

\begin{figure}[!ht]
\begin{center}
    \subfigure[Original image]{
    	\includegraphics[scale=0.115]{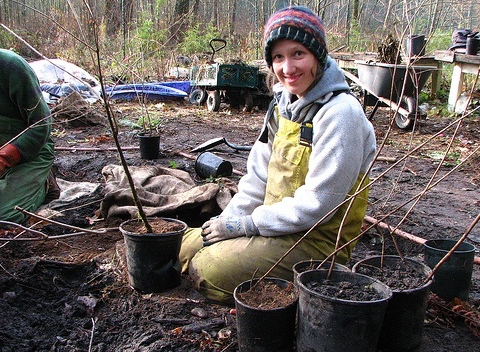}
    	\includegraphics[scale=0.115]{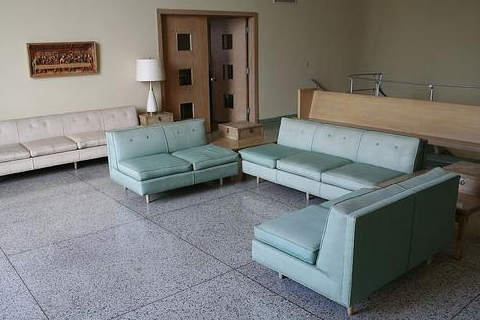}
    	\includegraphics[scale=0.115]{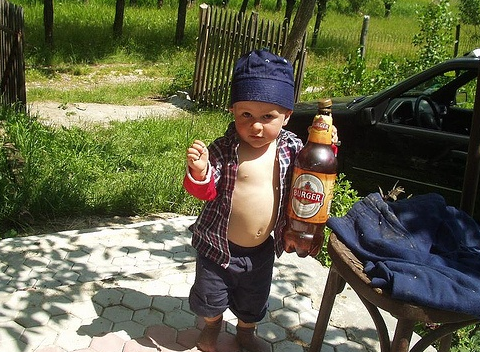}
    	\includegraphics[scale=0.115]{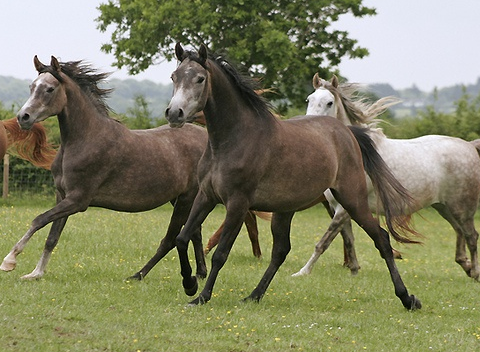}
    }
    \subfigure[Ground truth]{
    	\includegraphics[scale=0.115]{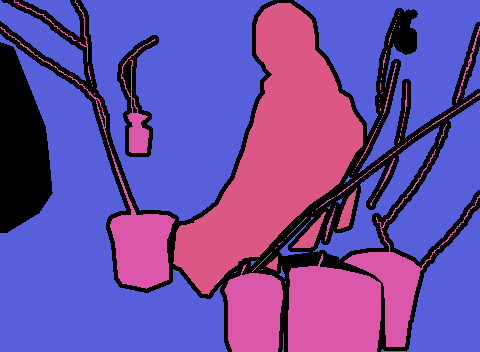}
    	\includegraphics[scale=0.115]{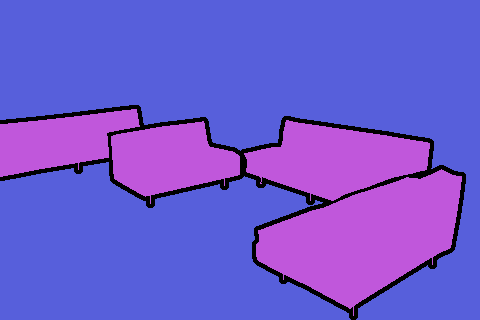}
    	\includegraphics[scale=0.115]{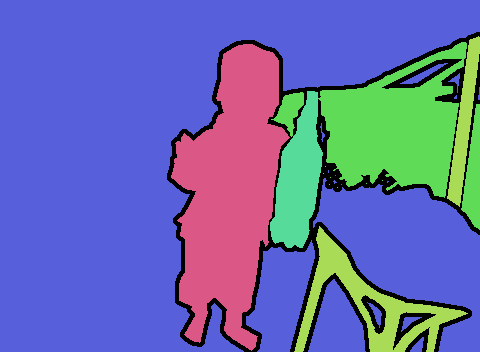}
    	\includegraphics[scale=0.115]{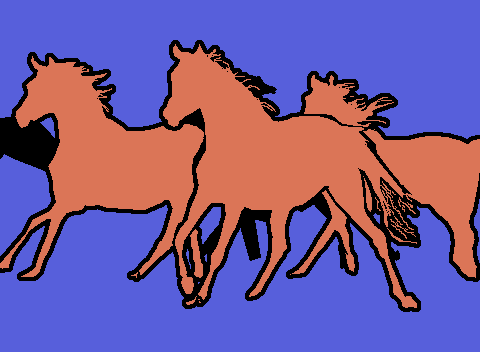}
    }
    \subfigure[32-bit FCN]{
    	\includegraphics[scale=0.115]{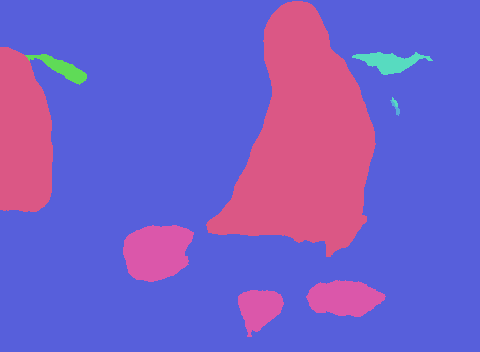}
    	\includegraphics[scale=0.115]{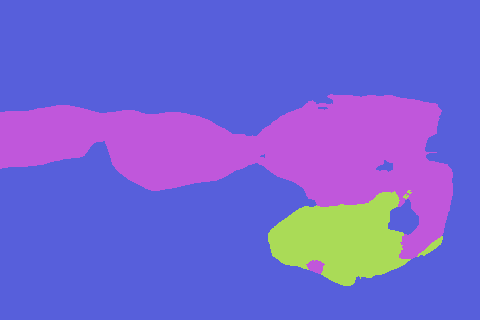}
    	\includegraphics[scale=0.115]{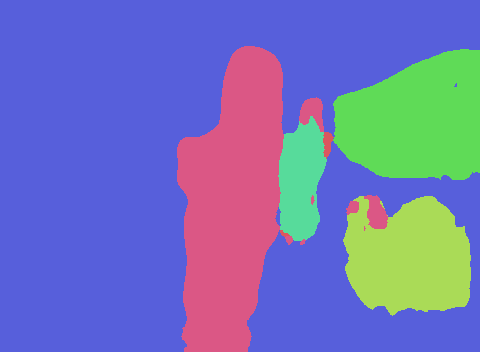}
    	\includegraphics[scale=0.115]{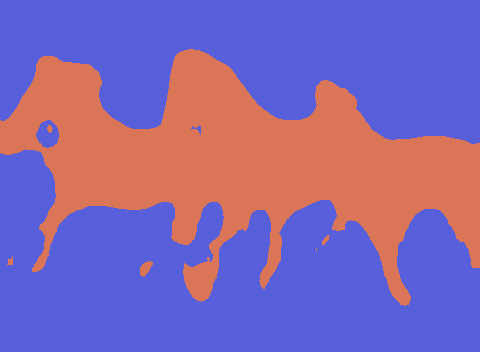}
    }
    \subfigure[2-bit BFCN]{
    	\includegraphics[scale=0.115]{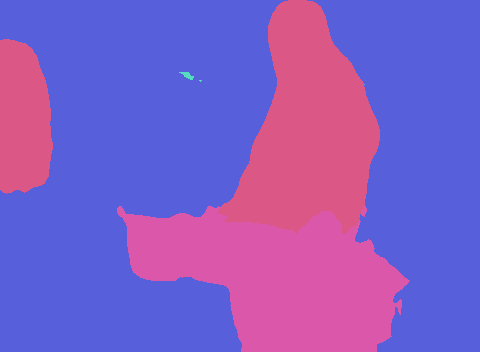}
    	\includegraphics[scale=0.115]{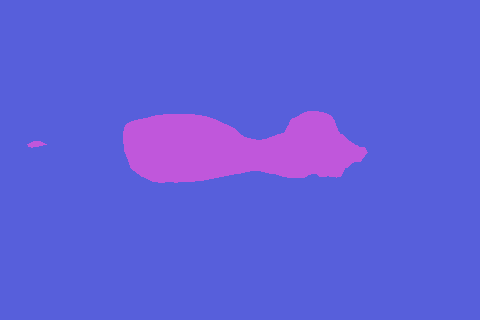}
    	\includegraphics[scale=0.115]{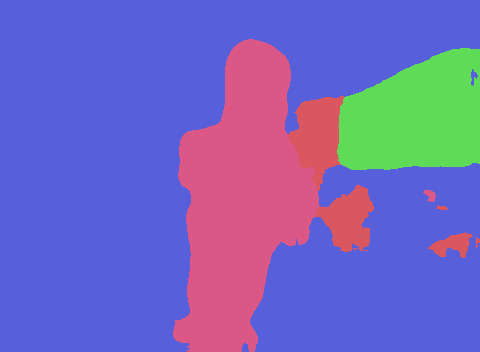}
    	\includegraphics[scale=0.115]{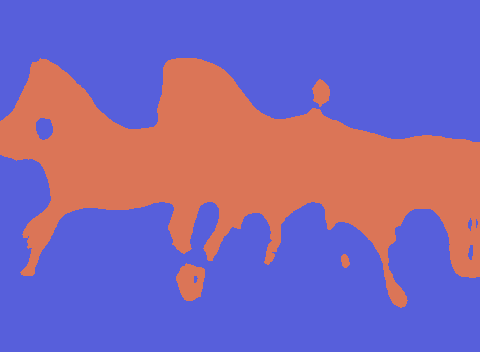}
    }
 
\end{center}
   \caption{Examples on PASCAL VOC 2012.}
\label{fig:examples}
\end{figure}

We present some example outputs on PASCAL VOC 2012 in Figure \ref{fig:examples}.
From predictions we can see that BFCNs perform well on easy tasks.
But on difficult tasks, which mostly consist of small objects or rare classes like bottle and sofa, BFCN will fail and have worse boundary performance.
It also seems that BFCN has difficulties in reconstructing fine structures of the input image.
However, low bit-width networks seldom misclassify the whole object, which effectively allow them to be used in real applications.

\section{Conclusion and Future Work}
In this paper, we propose and study methods for training bit fully convolutional network,
which uses low bit-width weights and activations to accelerate inference speed and reduce memory footprint.
We also propose a novel method to train a low bit-width network, which decreases bit-width step by step
to reduce performance loss resulting from quantization.
As a result, we are able to train efficient low bit-width scene-parsing networks
without losing much performance.
The low bit-width networks are especially friendly to hardware implementations like FPGA
as low bit-width multipliers usually require orders of magnitude less resources.

As future work, a better baseline model can be used and CRF as well as other techniques can be integrated into BFCN for even better performance.
We also note that our methods of designing and training low bit-width network can also be applied to other related tasks
such as object detection and instance segmentation.

\bibliographystyle{aaai}
\bibliography{thesis}

\begin{thebibliography}{}

\bibitem[\protect\citeauthoryear{Badrinarayanan, Handa, and
  Cipolla}{2015}]{badrinarayanan2015segnet}
Badrinarayanan, V.; Handa, A.; and Cipolla, R.
\newblock 2015.
\newblock Segnet: A deep convolutional encoder-decoder architecture for robust
  semantic pixel-wise labelling.
\newblock {\em arXiv preprint arXiv:1505.07293}.

\bibitem[\protect\citeauthoryear{Chen \bgroup et al\mbox.\egroup
  }{2014a}]{chen2014semantic}
Chen, L.-C.; Papandreou, G.; Kokkinos, I.; Murphy, K.; and Yuille, A.~L.
\newblock 2014a.
\newblock Semantic image segmentation with deep convolutional nets and fully
  connected crfs.
\newblock {\em arXiv preprint arXiv:1412.7062}.

\bibitem[\protect\citeauthoryear{Chen \bgroup et al\mbox.\egroup
  }{2014b}]{chen2014diannao}
Chen, T.; Du, Z.; Sun, N.; Wang, J.; Wu, C.; Chen, Y.; and Temam, O.
\newblock 2014b.
\newblock Diannao: A small-footprint high-throughput accelerator for ubiquitous
  machine-learning.
\newblock In {\em ACM Sigplan Notices}, volume~49,  269--284.
\newblock ACM.

\bibitem[\protect\citeauthoryear{Chen \bgroup et al\mbox.\egroup
  }{2014c}]{chen2014dadiannao}
Chen, Y.; Luo, T.; Liu, S.; Zhang, S.; He, L.; Wang, J.; Li, L.; Chen, T.; Xu,
  Z.; Sun, N.; et~al.
\newblock 2014c.
\newblock Dadiannao: A machine-learning supercomputer.
\newblock In {\em Microarchitecture (MICRO), 2014 47th Annual IEEE/ACM
  International Symposium on},  609--622.
\newblock IEEE.

\bibitem[\protect\citeauthoryear{Cordts \bgroup et al\mbox.\egroup
  }{2016}]{Cordts2016Cityscapes}
Cordts, M.; Omran, M.; Ramos, S.; Rehfeld, T.; Enzweiler, M.; Benenson, R.;
  Franke, U.; Roth, S.; and Schiele, B.
\newblock 2016.
\newblock The cityscapes dataset for semantic urban scene understanding.
\newblock In {\em Proc. of the IEEE Conference on Computer Vision and Pattern
  Recognition (CVPR)}.

\bibitem[\protect\citeauthoryear{Courbariaux, Bengio, and
  David}{2014}]{courbariaux2014training}
Courbariaux, M.; Bengio, Y.; and David, J.-P.
\newblock 2014.
\newblock Training deep neural networks with low precision multiplications.
\newblock {\em arXiv preprint arXiv:1412.7024}.

\bibitem[\protect\citeauthoryear{Everingham \bgroup et al\mbox.\egroup
  }{2015}]{Everingham15}
Everingham, M.; Eslami, S. M.~A.; Van~Gool, L.; Williams, C. K.~I.; Winn, J.;
  and Zisserman, A.
\newblock 2015.
\newblock The pascal visual object classes challenge: A retrospective.
\newblock {\em International Journal of Computer Vision} 111(1):98--136.

\bibitem[\protect\citeauthoryear{Farabet \bgroup et al\mbox.\egroup
  }{2009}]{farabet2009cnp}
Farabet, C.; Poulet, C.; Han, J.~Y.; and LeCun, Y.
\newblock 2009.
\newblock Cnp: An fpga-based processor for convolutional networks.
\newblock In {\em 2009 International Conference on Field Programmable Logic and
  Applications},  32--37.
\newblock IEEE.

\bibitem[\protect\citeauthoryear{Farabet \bgroup et al\mbox.\egroup
  }{2011}]{farabet2011large}
Farabet, C.; LeCun, Y.; Kavukcuoglu, K.; Culurciello, E.; Martini, B.;
  Akselrod, P.; and Talay, S.
\newblock 2011.
\newblock Large-scale fpga-based convolutional networks.
\newblock {\em Machine Learning on Very Large Data Sets} 1.

\bibitem[\protect\citeauthoryear{Ghiasi and Fowlkes}{2016}]{ghiasi16lrr}
Ghiasi, G., and Fowlkes, C.~C.
\newblock 2016.
\newblock Laplacian reconstruction and refinement for semantic segmentation.
\newblock {\em CoRR} abs/1605.02264.

\bibitem[\protect\citeauthoryear{Gong \bgroup et al\mbox.\egroup
  }{2014}]{gong2014compressing}
Gong, Y.; Liu, L.; Yang, M.; and Bourdev, L.
\newblock 2014.
\newblock Compressing deep convolutional networks using vector quantization.
\newblock {\em arXiv preprint arXiv:1412.6115}.

\bibitem[\protect\citeauthoryear{Gupta \bgroup et al\mbox.\egroup
  }{2015}]{gupta2015deep}
Gupta, S.; Agrawal, A.; Gopalakrishnan, K.; and Narayanan, P.
\newblock 2015.
\newblock Deep learning with limited numerical precision.
\newblock {\em arXiv preprint arXiv:1502.02551}.

\bibitem[\protect\citeauthoryear{Han \bgroup et al\mbox.\egroup
  }{2015}]{han2015learning}
Han, S.; Pool, J.; Tran, J.; and Dally, W.
\newblock 2015.
\newblock Learning both weights and connections for efficient neural network.
\newblock In {\em Advances in Neural Information Processing Systems},
  1135--1143.

\bibitem[\protect\citeauthoryear{Hariharan \bgroup et al\mbox.\egroup
  }{2011}]{hariharan2011semantic}
Hariharan, B.; Arbel{\'a}ez, P.; Bourdev, L.; Maji, S.; and Malik, J.
\newblock 2011.
\newblock Semantic contours from inverse detectors.
\newblock In {\em 2011 International Conference on Computer Vision},  991--998.
\newblock IEEE.

\bibitem[\protect\citeauthoryear{He \bgroup et al\mbox.\egroup
  }{2015}]{he2015deep}
He, K.; Zhang, X.; Ren, S.; and Sun, J.
\newblock 2015.
\newblock Deep residual learning for image recognition.
\newblock {\em arXiv preprint arXiv:1512.03385}.

\bibitem[\protect\citeauthoryear{Huang \bgroup et al\mbox.\egroup
  }{2015}]{huang2015densebox}
Huang, L.; Yang, Y.; Deng, Y.; and Yu, Y.
\newblock 2015.
\newblock Densebox: Unifying landmark localization with end to end object
  detection.
\newblock {\em arXiv preprint arXiv:1509.04874}.

\bibitem[\protect\citeauthoryear{Johnson, Karpathy, and
  Fei-Fei}{2015}]{johnson2015densecap}
Johnson, J.; Karpathy, A.; and Fei-Fei, L.
\newblock 2015.
\newblock Densecap: Fully convolutional localization networks for dense
  captioning.
\newblock {\em arXiv preprint arXiv:1511.07571}.

\bibitem[\protect\citeauthoryear{Kim and Smaragdis}{2016}]{kim2016bitwise}
Kim, M., and Smaragdis, P.
\newblock 2016.
\newblock Bitwise neural networks.
\newblock {\em arXiv preprint arXiv:1601.06071}.

\bibitem[\protect\citeauthoryear{Kim \bgroup et al\mbox.\egroup
  }{2016}]{kim2016pvanet}
Kim, K.-H.; Cheon, Y.; Hong, S.; Roh, B.; and Park, M.
\newblock 2016.
\newblock Pvanet: Deep but lightweight neural networks for real-time object
  detection.
\newblock {\em arXiv preprint arXiv:1608.08021}.

\bibitem[\protect\citeauthoryear{Krizhevsky, Sutskever, and
  Hinton}{2012}]{krizhevsky2012imagenet}
Krizhevsky, A.; Sutskever, I.; and Hinton, G.~E.
\newblock 2012.
\newblock Imagenet classification with deep convolutional neural networks.
\newblock In {\em Advances in neural information processing systems},
  1097--1105.

\bibitem[\protect\citeauthoryear{Long, Shelhamer, and
  Darrell}{2015}]{long2015fully}
Long, J.; Shelhamer, E.; and Darrell, T.
\newblock 2015.
\newblock Fully convolutional networks for semantic segmentation.
\newblock In {\em Proceedings of the IEEE Conference on Computer Vision and
  Pattern Recognition},  3431--3440.

\bibitem[\protect\citeauthoryear{Paszke \bgroup et al\mbox.\egroup
  }{2016}]{paszke2016enet}
Paszke, A.; Chaurasia, A.; Kim, S.; and Culurciello, E.
\newblock 2016.
\newblock Enet: A deep neural network architecture for real-time semantic
  segmentation.
\newblock {\em arXiv preprint arXiv:1606.02147}.

\bibitem[\protect\citeauthoryear{Pham \bgroup et al\mbox.\egroup
  }{2012}]{pham2012neuflow}
Pham, P.-H.; Jelaca, D.; Farabet, C.; Martini, B.; LeCun, Y.; and Culurciello,
  E.
\newblock 2012.
\newblock Neuflow: Dataflow vision processing system-on-a-chip.
\newblock In {\em Circuits and Systems (MWSCAS), 2012 IEEE 55th International
  Midwest Symposium on},  1044--1047.
\newblock IEEE.

\bibitem[\protect\citeauthoryear{Rastegari \bgroup et al\mbox.\egroup
  }{2016}]{rastegari2016xnor}
Rastegari, M.; Ordonez, V.; Redmon, J.; and Farhadi, A.
\newblock 2016.
\newblock Xnor-net: Imagenet classification using binary convolutional neural
  networks.
\newblock {\em arXiv preprint arXiv:1603.05279}.

\bibitem[\protect\citeauthoryear{Ren \bgroup et al\mbox.\egroup
  }{2015}]{ren2015faster}
Ren, S.; He, K.; Girshick, R.; and Sun, J.
\newblock 2015.
\newblock Faster r-cnn: Towards real-time object detection with region proposal
  networks.
\newblock In {\em Advances in neural information processing systems},  91--99.

\bibitem[\protect\citeauthoryear{Simonyan and Zisserman}{2014}]{Simonyan14c}
Simonyan, K., and Zisserman, A.
\newblock 2014.
\newblock Very deep convolutional networks for large-scale image recognition.
\newblock {\em CoRR} abs/1409.1556.

\bibitem[\protect\citeauthoryear{Vanhoucke, Senior, and
  Mao}{2011}]{vanhoucke2011improving}
Vanhoucke, V.; Senior, A.; and Mao, M.~Z.
\newblock 2011.
\newblock Improving the speed of neural networks on cpus.
\newblock In {\em Proc. Deep Learning and Unsupervised Feature Learning NIPS
  Workshop}, volume~1.

\bibitem[\protect\citeauthoryear{Wu, Shen, and Hengel}{2016}]{wu2016high}
Wu, Z.; Shen, C.; and Hengel, A. v.~d.
\newblock 2016.
\newblock High-performance semantic segmentation using very deep fully
  convolutional networks.
\newblock {\em arXiv preprint arXiv:1604.04339}.

\bibitem[\protect\citeauthoryear{Zhang \bgroup et al\mbox.\egroup
  }{2015a}]{zhang2015optimizing}
Zhang, C.; Li, P.; Sun, G.; Guan, Y.; Xiao, B.; and Cong, J.
\newblock 2015a.
\newblock Optimizing fpga-based accelerator design for deep convolutional
  neural networks.
\newblock In {\em Proceedings of the 2015 ACM/SIGDA International Symposium on
  Field-Programmable Gate Arrays},  161--170.
\newblock ACM.

\bibitem[\protect\citeauthoryear{Zhang \bgroup et al\mbox.\egroup
  }{2015b}]{zhang2015accelerating}
Zhang, X.; Zou, J.; He, K.; and Sun, J.
\newblock 2015b.
\newblock Accelerating very deep convolutional networks for classification and
  detection.

\bibitem[\protect\citeauthoryear{Zhou \bgroup et al\mbox.\egroup
  }{2016}]{zhou2016dorefa}
Zhou, S.; Wu, Y.; Ni, Z.; Zhou, X.; Wen, H.; and Zou, Y.
\newblock 2016.
\newblock Dorefa-net: Training low bitwidth convolutional neural networks with
  low bitwidth gradients.
\newblock {\em arXiv preprint arXiv:1606.06160}.

\end{thebibliography}

\end{document}